\documentclass[10pt,twocolumn,letterpaper]{article}
\usepackage{iccv}     
\usepackage{amsmath}
\usepackage{booktabs}
\usepackage{etoolbox,siunitx}
\usepackage{microtype}
\usepackage{multirow}
\usepackage{tabularx}
\usepackage{xr}

\definecolor{iccvblue}{rgb}{0.21,0.49,0.74}
\usepackage[pagebackref,breaklinks,colorlinks,allcolors=iccvblue]{hyperref}

\robustify\bfseries
\newcommand*{\inparagraph}[1]{\noindent\textbf{#1}\hspace{0.5em}}

\newcommand{\oursName}{BA-Track\@\xspace}
\newcommand{\row}[1]{\textit{(#1)}\@\xspace}

\DeclareMathOperator*{\argmin}{arg\,min}

\newcommand\blfootnote[1]{%
  \begingroup
  \renewcommand\thefootnote{}\footnote{#1}%
  \addtocounter{footnote}{-1}%
  \endgroup
}

\title{Back on Track: Bundle Adjustment for Dynamic Scene Reconstruction}
\author{
Weirong Chen\textsuperscript{1,2} \ \
Ganlin Zhang\textsuperscript{1,2} \ \
Felix Wimbauer\textsuperscript{1,2} \ \
Rui Wang\textsuperscript{4} \ \
Nikita Araslanov\textsuperscript{1,2} \ \ \\[2pt]
Andrea Vedaldi\textsuperscript{3} \quad \
Daniel Cremers\textsuperscript{1,2}\\[13pt]
{\textsuperscript{1}TU Munich\quad}
\,
{\textsuperscript{2}Munich Center for Machine Learning\quad}
\,
{\textsuperscript{3}University of Oxford\quad}
\,
{\textsuperscript{4}Microsoft}
\\[3pt]
}

\begin{document}
\maketitle
\begin{abstract}
Traditional SLAM systems, which rely on bundle adjustment, struggle with the highly dynamic scenes commonly found in casual videos. Such videos entangle the motion of dynamic elements, undermining the assumption of static environments required by traditional systems. Existing techniques either filter out dynamic elements or model their motion independently. However, the former often results in incomplete reconstructions, while the latter can lead to inconsistent motion estimates. Taking a novel approach, this work leverages a 3D point tracker to separate camera-induced motion from the observed motion of dynamic objects. By considering only the camera-induced component, bundle adjustment can operate reliably on all scene elements. We further ensure depth consistency across video frames with lightweight post-processing based on scale maps. Our framework combines the core of traditional SLAM---bundle adjustment---with a robust learning-based 3D tracker. Integrating motion decomposition, bundle adjustment, and depth refinement, our unified framework, BA-Track, accurately tracks camera motion and produces temporally coherent and scale-consistent dense reconstructions, accommodating both static and dynamic elements. Our experiments on challenging datasets reveal significant improvements in camera pose estimation and 3D reconstruction accuracy.
\blfootnote{Project page: \href{https://wrchen530.github.io/projects/batrack/}{wrchen530.github.io/projects/batrack}}
\end{abstract}    
\section{Introduction}%
\label{sec:intro}

\begin{figure}[t]
    \centering
    \includegraphics[width=\linewidth]{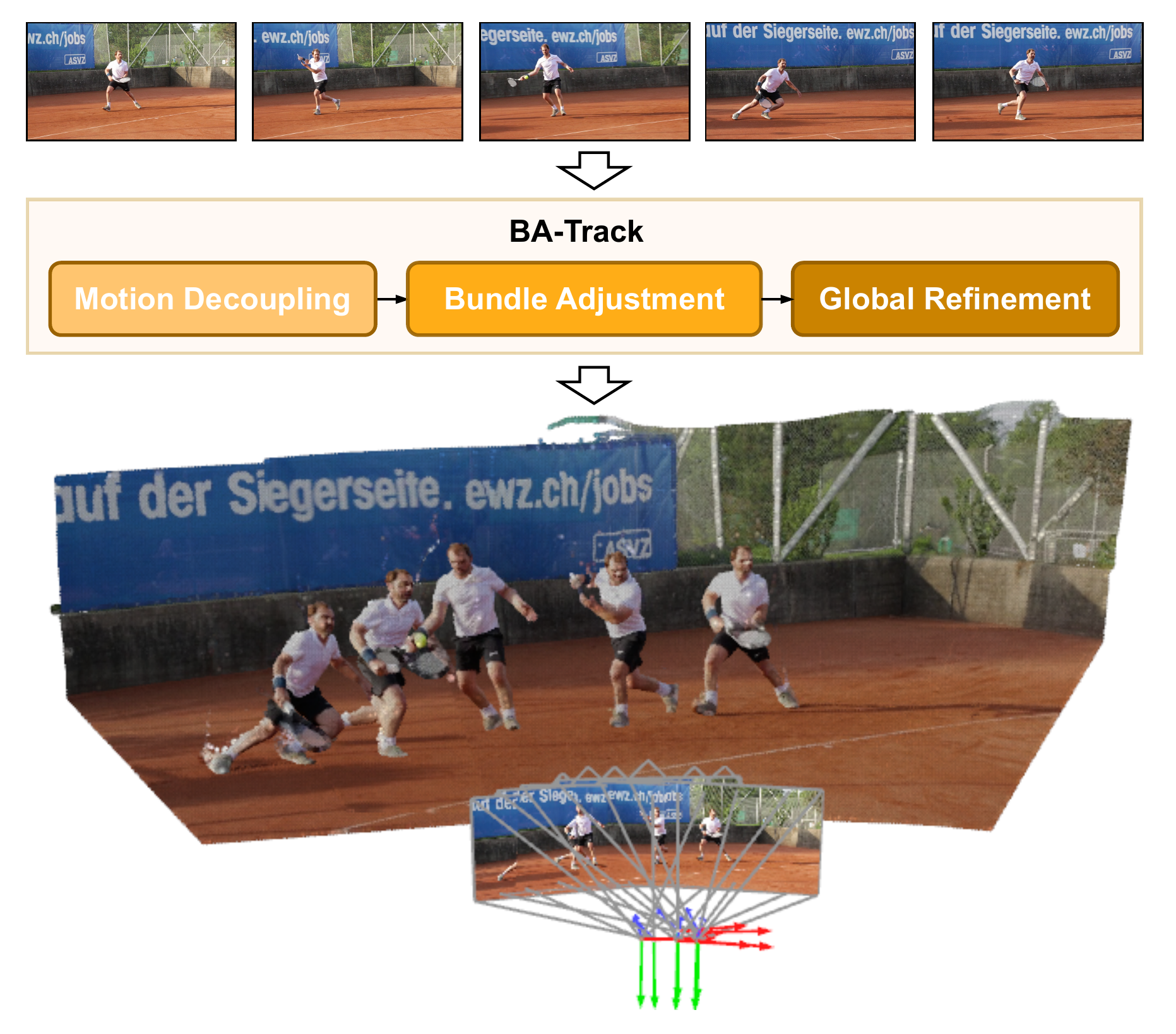}
    \caption{\textbf{Framework preview.} 
    Given a casual input video, \oursName uses a 3D tracker to separate camera-induced motion from the total observed motion, enabling bundle adjustment to process both static and dynamic points. Using the aligned sparse point tracks from bundle adjustment, we refine the dense depth maps, producing a globally consistent dynamic scene reconstruction.
    }%
    \label{fig:teaser}
\end{figure}

\begin{figure*}[ht]
    \centering
    \includegraphics[width=\linewidth]{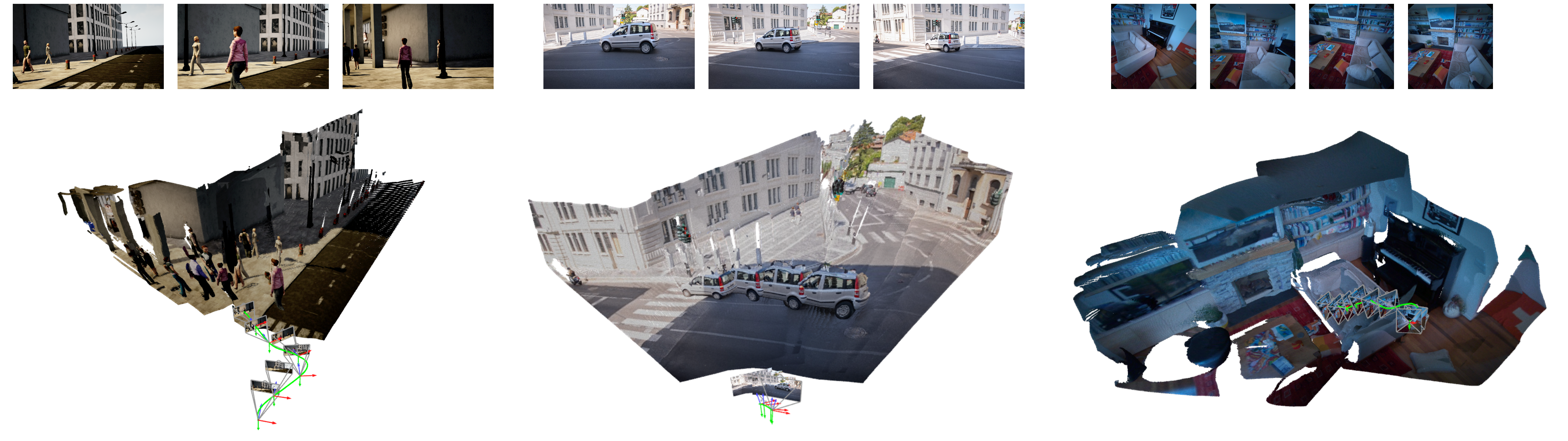}
    \vspace{-2em}
    \caption{\textbf{Dynamic scene reconstruction results on Shibuya~\cite{qiu2022airdos}, DAVIS~\cite{khoreva2019video}, and Aria Everyday Activities~\cite{lv2024aria} datasets.} By leveraging sparse dynamic SLAM and global refinement, \oursName achieves consistent dense 3D reconstructions across diverse dynamic scenes.}%
    \label{fig:3drecon}
    \vspace{-0.4em}
\end{figure*}

With the increasing prevalence of casual videos, reconstructing scenes containing moving objects has become essential for applications like augmented reality and robotics. Traditional methods for structure-from-motion (SfM)~\cite{Agarwal:2011:BRD} and simultaneous localization and mapping (SLAM)~\cite{Engel:2014:LSD} excel at recovering camera poses and scene geometry.
However, they leverage the epipolar constraint and thus apply only to static environments. 
Dynamic scenes with moving objects introduce ambiguities that violate the epipolar constraint. 
Resolving these ambiguities by analytical constraints or statistical outlier filtering alone is infeasible due to the complexity and diversity of real-world object motion.
This challenge motivates a hybrid approach that combines traditional optimization with deep priors.
Existing hybrid methods often address dynamic elements by detecting and removing them from optimization, leading to incomplete reconstructions or missing details associated with moving objects~\cite{bescos2018dynaslam, zhao2022particlesfm}.
Another approach is to rely on geometric cues, such as monocular depth models for per-frame reconstruction, but this requires additional optimization to achieve a globally aligned 3D reconstruction~\cite{kopf2021robust}.
This proves challenging in practice due to the inconsistency between estimated camera pose and depth priors across different frames, which can result in misaligned depth maps when integrating them into a coherent 3D scene.

We propose \oursName, a framework that jointly reconstructs static and dynamic scene geometry from casual video sequences (see \cref{fig:teaser}). Following traditional SfM and SLAM approaches, \oursName extends the time-honored bundle adjustment (BA)~\cite{Triggs:1999:BAA} to dynamic scene reconstruction.
The novel insight is to integrate a learning-based 3D tracker to decouple the motion of dynamic elements from the observed motion. 
By disentangling the camera-induced motion component of dynamic objects, \oursName reconstructs 3D trajectories of dynamic points as if they were stationary relative to their local reference frame.
As a result, the epipolar constraint becomes applicable to \emph{all points}, dynamic or static, enabling robust operation of bundle adjustment.

To effectively decouple motion, we incorporate monocular depth priors with a 3D tracker.
The tracker learns valuable 3D priors that help distinguish between camera- and object-based motion. 
Although such learning-based priors prove extremely helpful for reconstructing dynamic scenes, the scale of the depth maps exhibits temporal and spatial inconsistency.
To address this issue, we develop a global refinement module, which leverages the accurate but sparse geometry from BA to refine the dense depth maps.  
As illustrated in \cref{fig:3drecon}, \oursName achieves compelling reconstruction quality of both static and dynamic scene elements.

In summary, our framework consists of three components: \textit{(1)} a robust 3D tracking \emph{front-end}, which decouples camera-induced motion from observed (total) motion; \textit{(2)} a bundle adjustment \emph{back-end} for accurate pose and depth estimation, which leverages camera-induced motion for all points; and \textit{(3)} \emph{global refinement}, to achieve dense, scale-consistent depth.
Our learnable pipeline extends epipolar geometry to dynamic scene reconstruction, leveraging the power of established optimization techniques. Our experiments demonstrate significant improvements in camera motion estimation and 3D scene reconstruction, even in challenging scenes.

\section{Related Work}%
\label{sec:related_work}

\inparagraph{Point Tracking.}
Point tracking techniques estimate the 2D positions of one or more query points across a sequence of images.
Revisiting the idea of video representation as moving particles~\cite{sand2008particle}, PIPs~\cite{harley2022particle} takes a learning-based approach with a feed-forward network that iteratively refines point trajectories using multi-scale local features.
TAP-Net~\cite{doersch2022tap} takes a different direction by leveraging global correlations to directly compute trajectories, thereby eliminating the need for iterative refinement.
Subsequent research has explored various techniques to improve accuracy and efficiency, including inter-track relationships~\cite{karaev2023cotracker}, temporal refinements~\cite{doersch2023tapir}, nearest-neighbor interpolation~\cite{le2024dense}, and bidirectional 4D correlation~\cite{cho2025local}.
Moving beyond 2D tracking, recent methods such as SpatialTracker~\cite{xiao2024spatialtracker} and SceneTracker~\cite{bescos2018dynaslam} elevate the task into 3D space by incorporating depth priors.

Our approach extends 3D tracking by introducing a novel motion decoupling strategy that separates camera motion from object motion, enabling more effective integration with bundle adjustment.

\begin{figure*}[t]
\centering
\includegraphics[width=\linewidth]{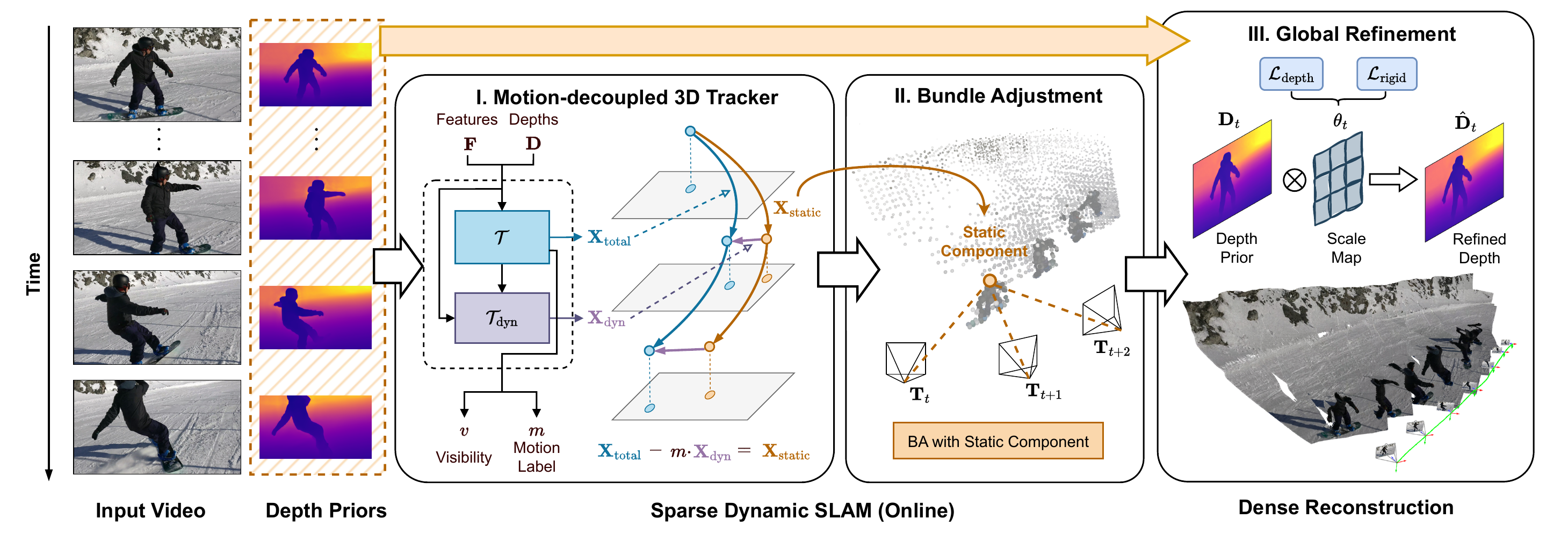}
\caption{\textbf{Overview of the \oursName framework.}
Given a temporal window, we compute image features $\mathbf{F}$ and depth features $\mathbf{D}$.
Our 3D tracker estimates local 3D tracks, visibility, dynamic labels, and decouples the static (camera-induced) motion of each query point.
Operating on the static motion components, bundle adjustment (BA) recovers the camera poses and global tracks.
The final refinement stage aligns the monocular depth priors with sparse BA estimates to ensure a temporally consistent and dense reconstruction.}%
\label{fig:pipeline}
\vspace{-0.4em}
\end{figure*}

\inparagraph{Monocular Visual Odometry.}
Monocular visual odometry (VO) estimates camera motion using RGB image sequences. Traditional methods can be divided into feature-based and direct approaches.
Feature-based methods detect and match keypoints across frames, using epipolar geometry and bundle adjustment for pose estimation~\cite{davison2007monoslam, mur2015orb}. Direct methods bypass keypoints, instead optimizing photometric consistency across frames using full-image information~\cite{Engel:2014:LSD, gao2018ldso, engel2017direct}.
Recent advances have shifted towards learning-based approaches that employ neural networks to enhance feature extraction and pose estimation~\cite{bloesch2018codeslam, teed2018deepv2d, teed2021droid, tang2018ba, wang2021tartanvo, teed2024deep, wimbauer2025anycam}.
For example, TartanVO~\cite{wang2021tartanvo} regresses motion from optical flow using deep networks, while DPVO~\cite{teed2024deep} integrates differentiable bundle adjustment with iterative correspondence updates.
Dynamic environments introduce additional challenges, as moving objects can interfere with motion estimation.
To improve robustness, some methods filter dynamic regions using masks~\cite{bescos2018dynaslam, shen2023dytanvo}, while others apply trajectory filtering to discard motion outliers~\cite{zhao2022particlesfm, Chen:2024:LEAP}.

\inparagraph{Dynamic Scene Reconstruction.}
Dynamic scene reconstruction aims to jointly recover geometry and motion in environments with moving objects.
Traditional methods such as SfM~\cite{schonberger2016structure} and MVS~\cite{schonberger2016pixelwise} assume static scenes, limiting their applicability in dynamic settings.
Regression-based approaches leverage large-scale training to estimate depth from single RGB images~\cite{birkl2023midas, bhat2023zoedepth, piccinelli2024unidepth, yang2024depthanyv2, ke2024repurposing}, while video-based extensions further improve temporal coherence by incorporating multi-frame cues, depth consistency, or diffusion priors~\cite{luo2020consistent, li2021enforcing, feng2022disentangling, hu2024depthcrafter}.
More recent approaches extend beyond depth to regress scene coordinates, modeling dynamic scenes through per-frame point maps~\cite{zhang2024monst3r, wang2025continuous, sucar2025dynamic, feng2025st4rtrack}.
To enforce geometric consistency, several methods combine learning with optimization, jointly refining scene structure and camera motion~\cite{kopf2021robust, zhang2022structure, zhang2024monst3r, li2025megasam}.
For example, MonST3R~\cite{zhang2024monst3r} uses a point map representation and refines global poses by filtering dynamic regions with an optical flow-based motion mask.
TracksTo4D~\cite{kasten2024fast}, closely related to our work, reconstructs dynamic scenes from 2D point trajectories by directly regressing point cloud bases and camera poses.
In contrast, our method decouples static and dynamic point motions and jointly optimizes them via bundle adjustment, achieving better robustness.
\section{BA-Track}%
\label{sec:method}

BA-Track consists of three stages (see~\cref{fig:pipeline}): \textit{(i)} A learning-based front-end decouples camera-induced motion from observed motion via a 3D tracker. 
\textit{(ii)} A bundle adjustment back-end recovers camera pose and sparse 3D geometry for both static and dynamic points using camera-induced motion. \textit{(iii)} A lightweight refinement stage leverages the sparse tracks to produce dense and temporally consistent depth.

\subsection{Stage I: Motion-Decoupled 3D Tracker}
\label{sec:tracker}

\begin{figure}[t]
  \centering
   \includegraphics[width=1.0\linewidth]{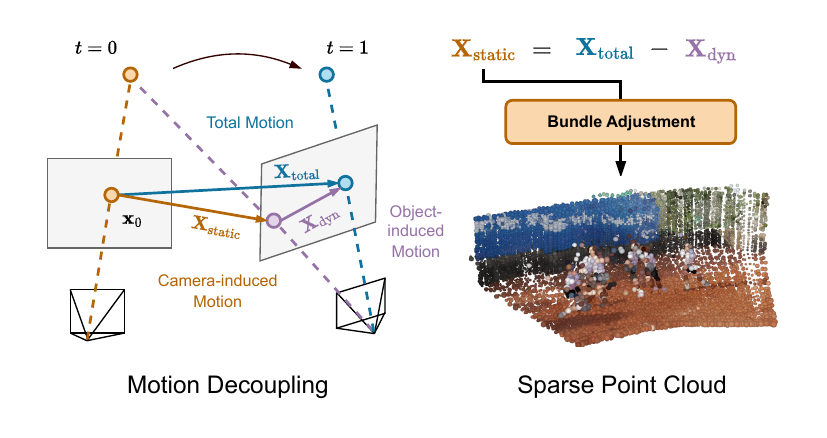}
    \vspace{-2em}
   \caption{\textbf{Illustration of motion decoupling.} We decompose the total observed point motion into a static component (induced by the camera motion) and a dynamic component (induced by object motion). The static component is then used by bundle adjustment to provide camera poses and sparse reconstruction.}%
   \label{fig:motion_decoup}
\end{figure}

Classical correspondence-based methods struggle to recover the 3D structure of dynamic objects from monocular videos due to violations of the epipolar constraint.
As a result, previous work requires detecting and filtering out dynamic regions to ensure that only static points are used for camera tracking and reconstruction~\cite{bescos2018dynaslam, zhao2022particlesfm, Chen:2024:LEAP}.  
Here, we explore an alternative approach.
Instead of discarding dynamic points, we infer the camera-induced component of their motion.
As shown in \cref{fig:motion_decoup}, the observed 2D motion comprises camera-induced (static) and object-induced (dynamic) components. 
By accurately estimating the static component, dynamic points become pseudo-static in their local reference frame.
Leveraging the static component, classical optimization, such as bundle adjustment, can operate seamlessly and requires no special treatment of dynamic points.

However, accurately estimating the camera-induced motion is non-trivial and requires strong motion priors.
To effectively learn motion priors, we integrate off-the-shelf depth cues and develop a 3D tracking front-end.
The tracker exploits the temporal context, while the additional depth cues enhance the tracker's 3D reasoning and improve its capacity to estimate the camera-induced motion.
Furthermore, we empirically observe that training a single network to estimate the camera-induced component is suboptimal. Instead, we employ two networks:
the first predicts the observed, ``total'' motion, while the other predicts the dynamic component corresponding to the object-induced motion.
We further discuss and empirically verify this design choice in \cref{sec:motion_decoupling}.

\inparagraph{Formulation.}
Considering a temporal window of RGB frames $\mathbf{I} = (I_1, \ldots, I_S)$ with resolution $H \times W$ and known camera intrinsics, we use a monocular depth estimation network~\cite{bhat2023zoedepth} to obtain their depth maps, $\mathbf{D} = (D_1, \ldots, D_S)$.
Given a 2D query point $\mathbf{x}^t = (u^t, v^t)$ in frame $t$, we sample its depth value $d^t = D_t[\mathbf{x}^t]$\footnote{$F[\mathbf{x}]$ denotes bilinear sampling from $F$ at location $\mathbf{x}$.} from the initial depth map, obtaining a 3D query point $X^t = (\mathbf{x}^t, D_t[\mathbf{x}^t])$. 
Since the intrinsics are known, we represent the 3D point using 2D coordinates along with the depth value.
The 3D tracker aims to estimate the corresponding 3D trajectory $\{ X^t(s) \mid s \in \{1, \dots, S\} \}$ for any timestep in the window.
For the remainder of this section, we omit the superscript $t$ indicating the query's frame of reference.

To extract relevant features for tracking, we pass each RGB image and its corresponding depth map through a CNN-based feature encoder~\cite{xiao2024spatialtracker}, and obtain $C$-dimensional feature maps, $\mathbf{F} = \{ F_1, \dots, F_S \}$, where $F_s \in \mathbb{R}^{C \times H/4 \times W/4}$.
We then define a point feature vector $f(s)$ associated with a 2D point $\mathbf{x}$, which encapsulates appearance, spatial, and depth information from the RGB-D input at each timestep $s$.

\inparagraph{Motion Decoupling.}
Our front-end consists of two transformer networks sharing the architecture with CoTracker~\cite{karaev2023cotracker}.
The tracker $\mathcal{T}$ predicts the observed motion $X_{\text{total}}$, visualized in blue in \cref{fig:pipeline}.
The dynamic tracker $\mathcal{T}_{\text{dyn}}$ is two times shallower than $\mathcal{T}$ for efficiency, and predicts the dynamic component $X_{\text{dyn}}$ of $X_{\text{total}}$.
The tracker $\mathcal{T}$ also learns point visibility $v \in [0, 1]^S$ and a static-dynamic label $m \in [0, 1]$.
Point visibility reflects occlusions, while the static-dynamic label indicates if the point belongs to a dynamic object.

After computing the total motion $X_{\text{total}}$ from the tracker $\mathcal{T}$, we use the dynamic tracker $\mathcal{T}_{\text{dyn}}$ to predict the dynamic motion component. 
For motion decoupling, we decompose $X_{\text{total}}$ into the camera motion $X_{\text{static}}$ (visualized in orange in \cref{fig:pipeline}) and the object motion $X_{\text{dyn}}$, weighted by the dynamic label $m$:
\begin{equation}\label{eq:decouple}
 X_{\text{static}} = X_{\text{total}} - m \cdot X_{\text{dyn}},  
\end{equation}
where $m$ acts as a gating factor.
Observe that the motion of static points ($m = 0$) coincides with the observed motion.
Conversely, the camera-induced motion of dynamic points ($m = 1$) results from subtracting their motion from the observed flow.
By decoupling the motion into static and dynamic parts, we can effectively track dynamic points as if they were static in their local frame of reference.
This novel mechanism allows the tracker to recover the 3D structure of dynamic objects with bundle adjustment in a seamless fashion, as we elaborate in \cref{sec:bundle_adj}.

\inparagraph{Training.} Under the hood, each transformer takes an initial 3D query location, extracts its point feature, and iteratively updates the point location and the corresponding point feature.
Each iteration leverages the local context of the query point, aggregated from the multi-scale feature correlations~\cite{karaev2023cotracker}.
We further incorporate depth map features into the context aggregation to embed 3D context.
Let us denote the point location and its feature vector after the $k$-th iteration as $(X^{(k)}, \: f^{(k)})$ for tracker $\mathcal{T}$, and $(X_{\text{dyn}}^{(k)}, f_{\text{dyn}}^{(k)})$ for the dynamic tracker $\mathcal{T}_{\text{dyn}}$.
Correspondingly to \cref{eq:decouple}, we obtain the static component for each iteration:
\begin{equation}
    X_{\text{static}}^{(k)} = X_{\text{total}}^{(k)} - m \cdot X_{\text{dyn}}^{(k)}.  
\end{equation}

We supervise $\mathcal{T}$ and $\mathcal{T}_\text{dyn}$ with ground-truth total 3D point trajectories and the static point trajectories from synthetic data.
To improve training convergence, we provide supervision for every iteration of the tracker,
\begin{equation}
    \mathcal{L}_{3D}(X^{(k)}) = \gamma^{K-k} \: \| X^{(k)} - X^\text{GT}\|_1,
\end{equation}
\begin{equation}
    \mathcal{L}_{3D}(X_{\text{static}}^{(k)}) = \gamma^{K-k} \: \| X_{\text{static}}^{(k)} - X_{\text{static}}^\text{GT}\|_1,
\end{equation}
where $K$ is the number of iterations and $\gamma$ is a hyperparameter (set empirically to $0.8$).

To learn the visibility and the static-dynamic label, we use binary cross-entropy, defined as
\begin{equation}
\begin{aligned}
    \mathcal{L}_{vis} &= (1-v^*) \log{(1-v)} + v^* \log{v}, \\
    \mathcal{L}_{dyn} &= (1-m^*) \log{(1-m)} + m^* \log{m},
\end{aligned}
\end{equation}
where $v^* \in \{ 0, 1\}^S$ and $m^\ast \in \{ 0, 1\}$ are the ground-truth labels.
Overall, the total loss for our 3D tracker is
\begin{equation}
\begin{aligned}
    \mathcal{L}_{total} = \mathcal{L}_{3D} + w_1 \mathcal{L}_{vis} + w_2 \mathcal{L}_{dyn},
\end{aligned}
\end{equation}
where $w_1$ and $w_2$ are scalar hyperparameters.

\subsection{Stage II: Bundle Adjustment}
\label{sec:bundle_adj}

We now employ our 3D tracker above to recover the camera poses and a sparse 3D structure of dynamic scenes.
Note that \cref{sec:tracker} exemplified tracking of a single query point, but our window-based 3D tracker can handle \emph{multiple} query points within a temporal window.
Here, we assume to have a set of 3D trajectories represented by their static 3D components $X_\text{static}$, as well as their per-frame visibility $v$ and the static-dynamic label $m$.
To simplify the notation, we use $X$ to denote the static component $X_\text{static}$ in the following.

Given a video sequence of length $L > S$, we extract a set of $N$ query points $\mathbf{X}^t = (X^t_1, \dots, X^t_N)$ from each frame $t \in \{1, \dots, L\}$.
Each query point $X^t_n = (u^t_n, v^t_n, d^t_n)$ represents the 2D pixel coordinates $\mathbf{x}^t_n = (u^t_n, v^t_n)$ of the point and its initial depth $d^t_n = D_t [\mathbf{x}^t_n]$.
Using our 3D tracker, we estimate a 3D trajectory of each query point $X^t_n$ within a local window of $S = 2S^\prime + 1$ frames.
Specifically, we obtain the trajectory as $\mathbf{X}_n^t = [X_n^t(1), \dots, X_n^t(S)]$ defined over the window \(I_{t-S^\prime:t+S^\prime}\).
Using a sliding window approach, we collect the 3D trajectories of all query points over the full sequence.
The result is a local 3D trajectory tensor \(\mathbf{X} \in \mathbb{R}^{L \times N \times S \times 3}\).

Since our tracker can estimate the static component of each point trajectory, we can leverage the classical BA optimization framework~\cite{Triggs:1999:BAA} for camera pose estimation and depth refinement.
This approach enhances the robustness of camera pose estimation, since it isolates the object motion and reduces the impact of dynamic points on pose accuracy.
In our BA framework, the optimization variables include the per-frame camera poses $\{\mathbf{T}_t \in SE(3) \}$ and the refined depths of the query points $\mathbf{Y} \in \mathbb{R}^{L \times N \times 1}$.
Each 3D trajectory $\mathbf{X}$ connects a query point $\mathbf{x}^i_n$ extracted from its source frame $i$ to its corresponding target locations in all other frames $j$.

The corresponding reprojection is defined as:
\begin{equation}
  \mathcal{P}_{j}(\mathbf{x}_{n}^i, y^i_{n}) = \Pi( \mathbf{T}_j \mathbf{T}_i^{-1} \Pi^{-1} (\mathbf{x}_{n}^i, y^i_{n})),  
\end{equation}
where $\Pi$ is the pinhole projection function and $\Pi^{-1}$ is the inverse projection. 
The RGB-D bundle adjustment is formulated as:

\begin{equation}
\begin{aligned}
    \argmin_{\{\mathbf{T}_t\}, \{\mathbf{Y}\}} \sum_{|i-j| \leq S} \sum_n  & W^i_{n}(j) \left \|\mathcal{P}_{j}(\mathbf{x}_{n}^i, y^i_{n}) - X_n^t(j)\right \|_\rho \\ +
    & \alpha \left \|y^i_{n} - d(\mathbf{X}^i_{n})\right \|^2,
\end{aligned}
\label{eq:ba}
\end{equation}
where $\left \|\cdot \right \|_\rho$ is the Huber loss and $\alpha$ balances the reprojection loss and depth consistency.
The confidence weight \(W^i_{j,n}\) encapsulates the visibility of point $\mathbf{x}_{n}^i$ in frame $j$ and its dynamic label by $ W^i_{n}(j) =  v^i_n(j) \cdot (1 - m_n^i)$.
We solve \cref{eq:ba} efficiently using the Gauss-Newton method with Schur decomposition~\cite{Triggs:1999:BAA}.

\begin{table*}[ht]
\centering
\small
\setlength{\abovecaptionskip}{0.1cm}
\setlength{\belowcaptionskip}{-0.15cm}
\setlength\tabcolsep{6pt}
\begin{tabularx}{\linewidth}{@{}lX@{\hspace{1em}}ccccccccccc@{}}
\toprule
\multirow{2}{*}{Category}    & \multirow{2}{*}{Method}                  & \multicolumn{3}{c}{MPI Sintel~\cite{Butler:2012:Sintel}} &                & \multicolumn{3}{c}{AirDOS Shibuya~\cite{qiu2022airdos}} &         & \multicolumn{3}{c}{Epic Fields~\cite{tschernezki2024epic}} \\ \cmidrule(lr){3-5} \cmidrule(lr){7-9} \cmidrule(lr){11-13}
                             &                                          & ATE            & RTE            & RRE            &         & ATE            & RTE            & RRE            &        & ATE              & RTE             & RRE            \\ \midrule
w/ intrinsics                & DROID-SLAM~\cite{teed2021droid}          & 0.175          & 0.084          & 1.912          &         & 0.256          & 0.123          & 0.628          &        & 1.424            & 0.130           & 2.180          \\
                             & TartanVO~\cite{wang2021tartanvo}         & 0.238          & 0.093          & 1.305          &         & 0.135          & 0.030          & 0.249          &        & 1.490            & 0.121           & 1.548          \\
                             & DytanVO~\cite{shen2023dytanvo}           & 0.131          & 0.097          & 1.538          &         & 0.088          & 0.026          & 0.251          &        & 1.608            & 0.119           & 1.556          \\
                             & DPVO~\cite{teed2024deep}                 & 0.115          & 0.072          & 1.975          &         & 0.146          & 0.091          & 0.387          &        & 0.394            & 0.078           & \textbf{0.734} \\
                             & LEAP-VO~\cite{Chen:2024:LEAP}            & 0.089          & 0.066          & 1.250          &         & 0.031          & 0.111          & \textbf{0.124} &        & 0.486            & 0.071           & 1.018          \\ \midrule
w/o intrinsics               & Robust-CVD~\cite{kopf2021robust}         & 0.360          & 0.154          & 3.443          &         & ---            & ---            & ---            &        & ---              & ---             & ---            \\
                             & ParticleSfM~\cite{zhao2022particlesfm}   & 0.129          & 0.031          & 0.535          &         & 0.275          & 0.155          & 0.750          &        & ---              & ---             & ---            \\
                             & CasualSAM~\cite{zhang2022structure}      & 0.141          & 0.035          & 0.615          &         & 0.209          & 0.202          & 0.620          &        & ---              & ---             & ---            \\                          
                             & MonST3R~\cite{piccinelli2024unidepth}    & 0.108          & 0.042          & 0.732          &         & (0.512)        & (0.075)        & (0.566)        &        & ---              & ---             & ---            \\
                             \midrule   w/ intrinsics
                             & \oursName (Ours)                         & \textbf{0.034} & \textbf{0.023} & \textbf{0.115} &         & \textbf{0.028} & \textbf{0.009} & 0.150          &        & \textbf{ 0.385 } & \textbf{ 0.066} & 1.029          \\ \bottomrule
\end{tabularx}
\caption{\textbf{Camera pose evaluation results on Sintel~\cite{Butler:2012:Sintel}, Shibuya~\cite{qiu2022airdos}, and Epic Fields~\cite{tschernezki2024epic} datasets.} $(\cdot)$ denotes evaluation on sub-sequence due to memory constraints. Our method shows better results on ATE compared to other competitive baselines.}%
\label{tab:cam_pose_eval}
\vspace{0.5em}
\end{table*}

\subsection{Stage III: Global Refinement}

Bundle adjustment recovers the camera poses and adjusts the depths of the query points in their source frames of reference.
However, BA processes only a sparse set of query points, so the 3D positions of the other points in the depth maps remain unaffected.
To ensure consistency of the depth maps with the sparse set of accurate 3D tracks estimated by BA, we need global refinement.
Towards this goal, we introduce a function $\mathcal{H}_\theta: D \rightarrow \hat{D}$, where $\hat{D}$ is the refined depth map and $\theta$ is the parameter set. 
Global refinement jointly optimizes the function parameters $\theta$ and sparse 3D trajectories to achieve consistent geometry.

Our approach enforces two types of consistency: depth consistency and scene rigidity. The depth consistency loss encourages the refined depth map to match the sparse 3D trajectories within each frame, ensuring coherent alignment of sparse and dense depths. The scene rigidity loss maintains relative 3D distances between static trajectories across frames, preserving the rigidity of static structures.

Inspired by previous work~\cite{kopf2021robust}, we define a 2D scale grid of resolution $H_g \times W_g$ for each frame, $\theta_t \in \mathbb{R}^{H_g \times W_g}$.
The grid has a coarser resolution compared to the original image resolution $H \times W$.
Its goal is to scale the depth estimates $D_t$: For each 2D query point $\mathbf{x}$ in frame $t$, we determine the associated depth scale using bilinear sampling on the grid $\theta_t$.
The refined depth is then
\begin{equation}
   \hat{D}_t[\mathbf{x}] = \theta_t[\mathbf{x}] \cdot D_t[\mathbf{x}], 
\end{equation}
where $\theta_t[\mathbf{x}]$ serves as a scaling factor to adjust the initial depth $D_t[\mathbf{x}]$. We further introduce a local scale variable $\sigma^t_n$ for each 3D point $X_n^t$ from the local 3D trajectory tensor $\mathbf{X}$, allowing us to refine sparse trajectories alongside the adjustments in the dense depth map. This combination of global and local refinements accounts for both large-scale depth variations and fine-grained details.
The depth consistency loss optimizes $\theta$ and $\sigma$ via:
\begin{equation}\label{eq:loss_depth}
    \mathcal{L}_{\text{depth}} = \sum_{t,n} \left \| \theta_t[\mathbf{x}^t_n] \cdot D_t[\mathbf{x}^t_n] - \sigma^t_n d^t_n \right \|,
\end{equation}
while the scene rigidity loss optimizes $\theta$ as:

\begin{equation}
\begin{aligned}\label{eq:loss_rigid}
\mathcal{L}_{\text{rigid}} = \sum_{|i-j| < S} \sum_{(a,b) \in N} W_{\text{static}} \big(&\left \|  P^i_a(j) -  P^i_b(j) \right \| - \\
&\left \| P^i_a -  P^i_b\right \|\big).
\end{aligned}
\end{equation}
Here,
\begin{equation}
    P^i_k(j) = \Pi^{-1}(\mathbf{x}_k^i, \hat{D}_i[\mathbf{x}_k^i]), \quad k\in \{a,b\},
\end{equation}
is the 3D position derived from back-projecting 2D points using the refined dense depth.
The weight $W_{\text{static}}$ filters out the dynamic points as $W_{\text{static}} = (1-m^i_a) \cdot (1-m^i_b)$.
Intuitively, $\mathcal{L}_{\text{rigid}}$ enforces constant distances between arbitrary static points $a$ and $b$ in the target frame $j$ and the source frame $i$.
We optimize the variables using the Adam optimizer~\cite{kingma2014adam}.
\begin{figure*}[ht]
    \centering
    \includegraphics[width=\linewidth]{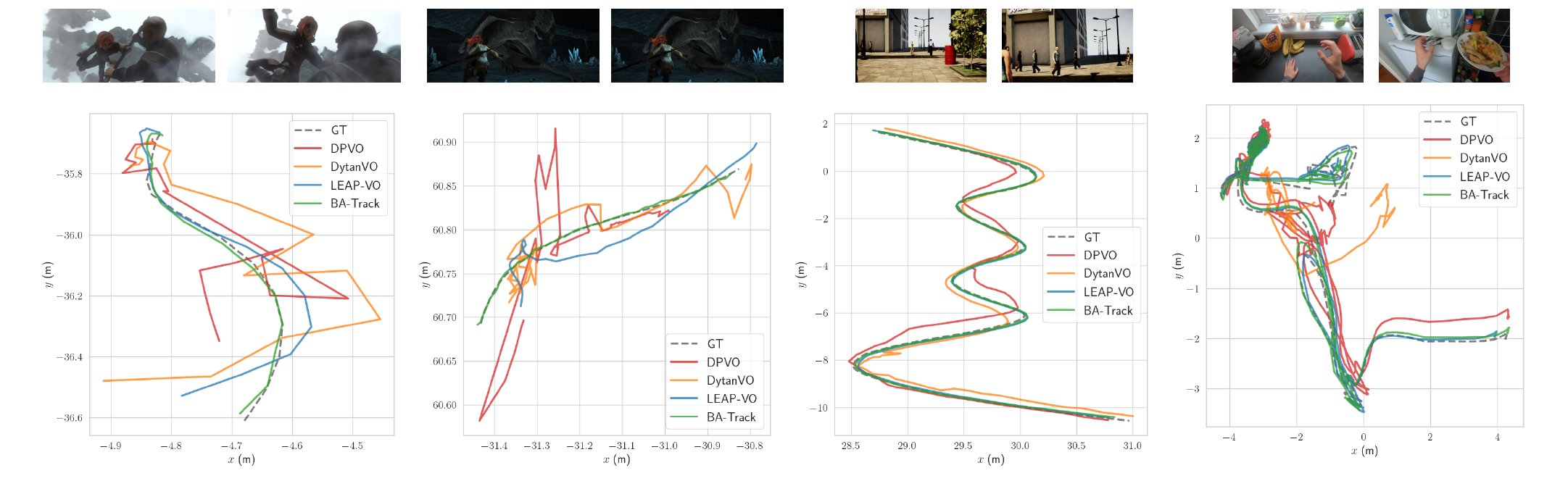}
    \caption{\textbf{Qualitative camera pose estimation results on Sintel~\cite{Butler:2012:Sintel}, Shibuya~\cite{qiu2022airdos}, and Epic Fields~\cite{tschernezki2024epic}.}
    Visualizations demonstrate that our method achieves more robust and accurate camera trajectories in challenging dynamic scenes.}%
    \label{fig:vis_cam_traj}
\end{figure*}

\section{Experiments}
\subsection{Implementation Details}
We train our 3D tracker on the TAP-Vid-Kubric training set~\cite{doersch2022tap}, which consists of 11,000 sequences, each with 24 frames. The training process uses the AdamW optimizer~\cite{Loshchilov:2019:DWD} with a learning rate of $3 \times 10^{-4}$ for a total of 100,000 steps, performed on 4 NVIDIA A100 GPUs. During training, we set the weights for visibility and dynamic labels to $w_1=5$ and $w_2=5$, employ $K=4$ iterative updates, a window size of $S=12$, and a total of $N=512$ query points. For BA, we use a window of 15 frames and apply 4 Gauss-Newton updates per window. We set $\alpha=0.05$ to balance reprojection and depth loss in \cref{eq:ba}. For BA weight filtering, we set $\delta_v = 0.9$ and $\delta_m = 0.9$. 
For further implementation details, please refer to our supplementary material.

\subsection{Camera Pose Evaluation}

\inparagraph{Datasets.}
We use three datasets for camera pose evaluation. 
MPI Sintel~\cite{Butler:2012:Sintel} provides sequences from 3D animated films featuring complex, fast object motion and exposure effects like motion blur. We use MPI Sintel to evaluate dynamic VO performance, with sequences spanning 20 to 50 frames.
AirDOS Shibuya~\cite{qiu2022airdos} includes sequences of 100 frames tracking over 30 individuals in two scene types. ``Road Crossing'' features diverse object motion, while ``Standing Humans'' contains dominant camera motion. EPIC Fields~\cite{tschernezki2024epic} includes egocentric videos with dynamic interactions, such as moving hands with kitchen tools, and is therefore ideal for studying long dynamic scenarios. We randomly select 8 sequences with reliable ground-truth trajectories, sampling every third frame of the first 3000 frames, yielding 1000 frames per sequence.

\begin{table*}[ht]
\centering
\small
\setlength\tabcolsep{7pt}
\begin{tabularx}{\linewidth}{@{}Xcccccccc@{}}
\toprule
\multirow{2}{*}{Method}    & \multicolumn{2}{c}{MPI Sintel~\cite{Butler:2012:Sintel}} & & \multicolumn{2}{c}{AirDOS Shibuya~\cite{qiu2022airdos}} & & \multicolumn{2}{c}{Bonn~\cite{palazzolo2019iros}} \\ \cmidrule(lr){2-3} \cmidrule(lr){5-6} \cmidrule(lr){8-9}
                          & Abs Rel   $\downarrow$      & $\delta < 1.25 \uparrow$        & & Abs Rel   $\downarrow$         & $\delta < 1.25 \uparrow$           & & Abs Rel   $\downarrow$    & $\delta < 1.25 \uparrow$      \\ \midrule
Robust-CVD~\cite{kopf2021robust}              & 0.703 & 47.8 &      & ---     & ---    &     & ---   & ---  \\
CasualSAM$^\dagger$~\cite{zhang2022structure} & 0.387 & 54.7 &      & 0.261   & 79.9   &     & 0.169 & 73.7 \\
MonST3R$^\ast$~\cite{ke2024repurposing}       & 0.335 & 58.5 &      & (0.208) & (71.2) &     & 0.063 & 96.4 \\ \midrule
ZoeDepth~\cite{bhat2023zoedepth}              & 0.467 & 47.3 &      & 0.571   & 43.8   &     & 0.087 & 94.8 \\
BA-Track with ZoeDepth (Ours)                 & 0.408 & 54.1 &      & 0.299   & 55.1   &     & 0.084 & 95.0 \\ \bottomrule
\end{tabularx}
\raggedright
$^\dagger$ test-time network finetuning; $^\ast$ depth from image pairs; the numbers in $(\cdot)$ denote results on a sub-sequence due to memory constraints.
\vspace{-0.05em}
\caption{\textbf{Depth evaluation results on Sintel~\cite{Butler:2012:Sintel}, Shibuya~\cite{qiu2022airdos}, and Bonn~\cite{palazzolo2019iros} datasets.} Our method achieves improved depth accuracy compared to the depth priors from ZoeDepth, demonstrating the effectiveness of our depth refinement.}%
\label{tab:depth_eval}
\vspace{-0.5em}
\end{table*}

\inparagraph{Metrics.}
We adopt three metrics. Absolute Translation Error (ATE) calculates the root mean square error between the estimated and ground-truth trajectories. Relative Translation Error (RTE) and Relative Rotation Error (RRE) measure translation (in meters) and rotation (in degrees) errors over a set distance, with both metrics averaged across all poses.
We perform all evaluations after $\operatorname{Sim}(3)$ Umeyama alignment~\cite{umeyama1991least} with the ground-truth trajectory.

\inparagraph{Results.}
We compare \oursName against learning-based SLAM and VO approaches, such as DROID-SLAM~\cite{teed2021droid}, ParticleSfM~\cite{zhao2022particlesfm},
DytanVO~\cite{shen2023dytanvo}, 
and LEAP-VO~\cite{Chen:2024:LEAP}.
Additionally, we compare \oursName to test-time optimization techniques, such as Robust-CVD~\cite{kopf2021robust} and CasualSAM~\cite{zhang2022structure}, 
which typically require several hours of processing time.

\Cref{tab:cam_pose_eval} shows a quantitative comparison of our method against state-of-the-art approaches across three benchmarks. Our approach consistently delivers superior performance in ATE and remains competitive in RTE and RRE\@.
On the MPI Sintel dataset, our method significantly outperforms all baselines. 
As \cref{fig:vis_cam_traj} illustrates, \oursName produces notably more accurate trajectory estimates, particularly in dynamic scenes where other methods struggle. On the challenging Epic Fields dataset, characterized by rapid dynamic content and complex camera motion, \oursName maintains compelling VO accuracy. 
Overall, \oursName demonstrates remarkable accuracy across all benchmarks with high memory efficiency. 
By contrast, prior work like MonST3R can only process up to 90 frames at once on a 48GB GPU\@.

\subsection{Depth Evaluation}

\inparagraph{Datasets.}
MPI Sintel~\cite{Butler:2012:Sintel} and AirDOS Shibuya~\cite{qiu2022airdos} are selected as both provide dense ground-truth depth maps.
We add a real-world dataset, Bonn RGB-D, which captures indoor activities such as object manipulation across 24 dynamic sequences. 
Following~\citet{zhang2024monst3r}, we select 5 dynamic sequences, each consisting of 110 frames.

\inparagraph{Metrics.}  
We adopt the absolute relative error (Abs Rel) and the threshold accuracy (TA) $\delta < 1.25$. Let $d_i$ and $d_i^*$ denote the predicted and the ground-truth depths for pixel $i$. Abs Rel is 
$\frac{1}{N} \sum_{i=1}^{N} \frac{|d_i^* - d_i|}{d_i^*}$ with $N$ the total number of pixels. TA is computed as the percentage of pixels satisfying $\max\left(\frac{d_i^*}{d_i}, \frac{d_i}{d_i^*}\right) < 1.25$.
We apply shift-scale alignment by computing a single shift and scale factor over the entire video to align the predicted depths with the ground truth~\cite{zhang2024monst3r}.

\begin{figure}[t]
    \centering
    \includegraphics[width=\linewidth]{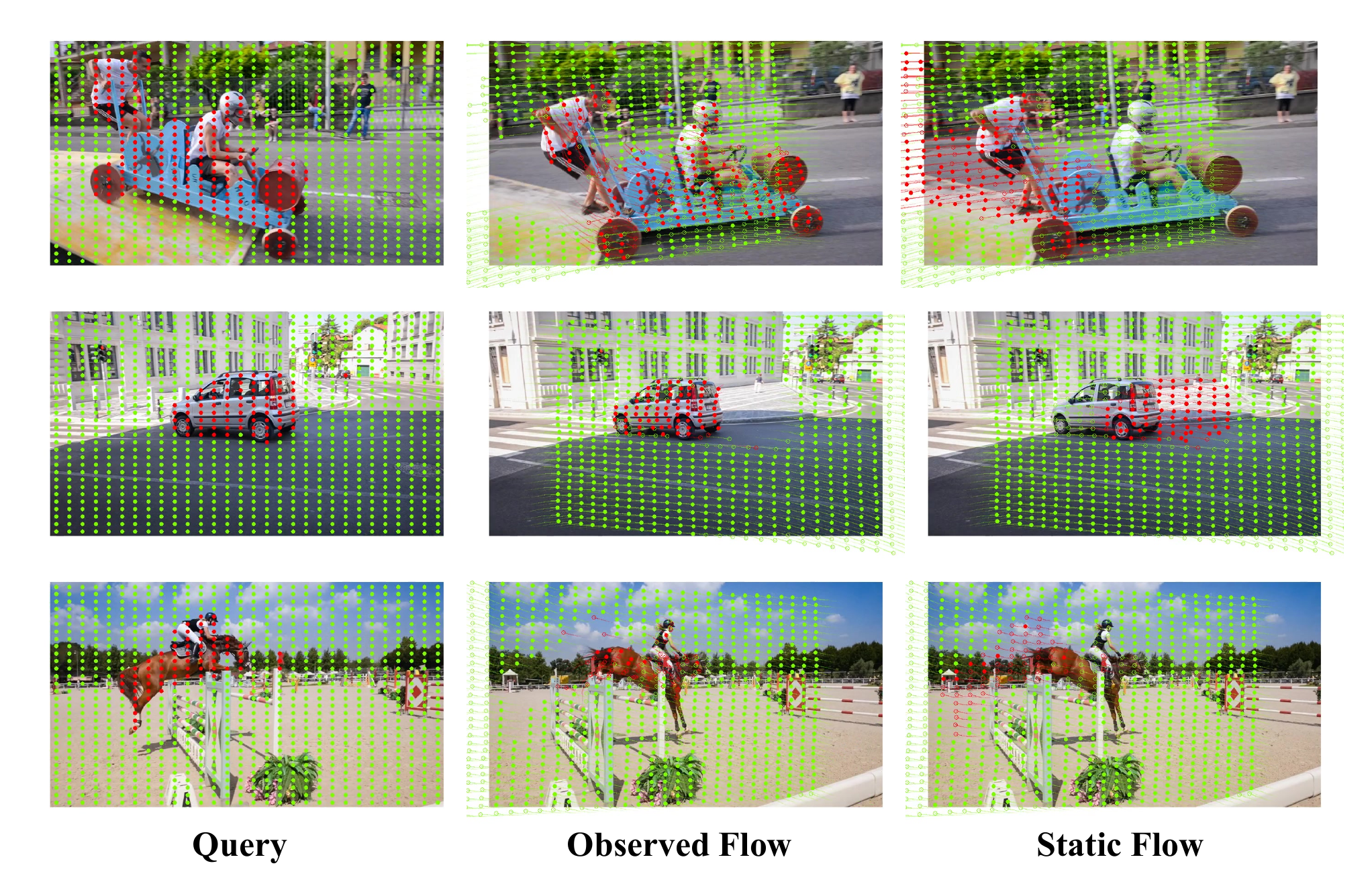}
    \vspace{-2em}
    \caption{
    \textbf{Motion decoupling on the DAVIS dataset~\cite{khoreva2019video}. }
   From left to right, the three columns show the reference frame for queries, total observed motion, and the static component. Green indicates static trajectories, and red indicates dynamic ones. The static component retains points on dynamic objects in their original (static) positions within the corresponding reference frame.
    }%
    \label{fig:motion_decouple}
    \vspace{-0.5em}
\end{figure}

\inparagraph{Results.} 
We compare \oursName to joint pose-depth optimization methods in \cref{tab:depth_eval}. 
Using ZoeDepth~\cite{bhat2023zoedepth} as input, our method consistently enhances depth estimates and achieves competitive accuracy compared to other joint refinement methods.
As we show later in \cref{sec:dynamic_reconstruction}, our global refinement plays a crucial role here, allowing \oursName to handle challenging scenarios with dynamic content. On MPI Sintel, we achieve an Abs Rel of 0.408 and an accuracy of 54.1\%, demonstrating comparable depth quality \wrt leading methods trained on much larger datasets, such as 
MonST3R~\cite{zhang2024monst3r}.
In more complex scenarios, such as the outdoor urban scenes in AirDOS Shibuya, our method maintains robust results with an Abs Rel of 0.299 and 55.1\% accuracy, outperforming competitors such as RobustCVD~\cite{kopf2021robust}. Our scale-map-based refinement offers efficient optimization with few parameters while remaining relatively simple. Investigating more advanced deformation models, such as neural networks, could be an interesting direction for future work.

\subsection{Motion Decoupling}%
\label{sec:motion_decoupling}

In our 3D tracker, motion decoupling isolates the static (camera-induced) component from the observed point motion, allowing us to treat dynamic points as if they were stationary. This approach reduces the impact of dynamic regions on camera pose estimation and dramatically enhances the robustness of BA\@.
It also aids the subsequent depth refinement, providing more coherent reconstructions of both static and dynamic parts.
\cref{fig:motion_decouple} illustrates the observed and static point motion on the DAVIS~\cite{khoreva2019video} dataset. Red points indicate dynamic trajectories, while green points represent static ones, highlighting the 3D tracker's ability to decouple the dynamic component from the observed motion.

To further demonstrate the benefits of motion decoupling, we analyze alternative motion representations for BA optimization in \cref{tab:pose_ablation}.
By applying motion decoupling with the dynamic tracker $\mathcal{T}_\text{dyn}$, we achieve a dramatic performance boost: The ATE drops by nearly half, from 0.137 \row{a} to 0.065 \row{e}.
We observe a further moderate boost in VO accuracy by setting the weights of dynamic points in BA optimization to zero \row{f}.
Note that this surpasses the VO accuracy of the setting with a single tracker predicting the observed motion \row{b}.
As an alternative tracker architecture, we investigate an approach with a static tracker (\textit{c-d}).
In this case, we train the tracker to directly regress the static component. However, this approach leads to inferior VO accuracy compared to our decoupling approach (\textit{e-f}).
Presumably, a single network struggles to learn two aspects simultaneously: visual tracking and motion patterns.
In contrast, our dual-network design addresses these tasks independently.

\subsection{Dynamic Reconstruction}%
\label{sec:dynamic_reconstruction}

\begin{table}[t]
\centering
\small
\setlength{\abovecaptionskip}{0.1cm}
\setlength{\belowcaptionskip}{-0.15cm}
\setlength\tabcolsep{4pt}
\begin{tabularx}{\linewidth}{@{}Xccccc@{}}
\toprule
\multirow{2}{*}{Setting} & \multicolumn{2}{c}{Dynamic Handling}        & \multicolumn{3}{c}{MPI Sintel}                   \\ 
\cmidrule(lr){2-3}  \cmidrule(lr){4-6} 
                         & Trajectory    & Camera Mask & ATE            & RTE            & RRE            \\ \midrule
(a)                      & Total         & ---         & 0.137          & 0.044          & 0.385          \\
(b)                      & Total         & \checkmark  & 0.047          & 0.025          & 0.137          \\ \midrule
(c)                      & Static*       & ---         & 0.091          & 0.038          & 0.380          \\
(d)                      & Static*       & \checkmark  & 0.072          & 0.032          & 0.233          \\ \midrule
(e)                      & Total-Dynamic & ---         & 0.065          & 0.025          & 0.197          \\
(f)                      & Total-Dynamic & \checkmark  & \textbf{0.034} & \textbf{0.023} & \textbf{0.115} \\ \bottomrule
\end{tabularx}
\caption{\textbf{Ablation study of dynamic handling methods on Sintel~\cite{Butler:2012:Sintel}}.
Combining motion decoupling with camera masking achieves the best results. $*$ denotes a setting trained for ablation.}%
\label{tab:pose_ablation}
\vspace{-0.0em}
\end{table}

\begin{table}[t]
\centering
\small
\begin{tabularx}{\linewidth}{@{}Xcc@{}S[table-format=1.3]@{\hspace{1em}}S[table-format=2.1]@{}}
\toprule
 \multirow{2}{*}{Method} & \multicolumn{2}{c}{Depth Refinement} & \multicolumn{2}{c}{Bonn crowd2~\cite{palazzolo2019iros}}       \\
 \cmidrule(lr){2-3} \cmidrule(lr){4-5} 
                           & {$\mathcal{L}_{\text{depth}}$} & {$\mathcal{L}_{\text{rigid}}$} & {Abs Rel $\downarrow$} & {$\delta < 1.25 \uparrow$} \\ \midrule
\multirow{4}{*}{\oursName} & ---                            & ---                            & 0.121                  & 89.6                       \\
                           & \checkmark                     & ---                            & 0.103                  & 94.8                       \\
                           & ---                            & \checkmark                     & 0.117                  & 88.4                       \\
                           & \checkmark                     & \checkmark                     & \bfseries 0.089        & \bfseries 95.0             \\ \bottomrule
\end{tabularx}
\caption{\textbf{Ablation study on depth refinement on Bonn crowd2 sequence~\cite{palazzolo2019iros}.} Applying both losses has a complementary effect.}%
\label{tab:depth_ablation}
\vspace{-0.8em}
\end{table}

We present qualitative results for camera tracking and 3D reconstruction on dynamic scenes in \cref{fig:3drecon} using an alternative depth model from UniDepth~\cite{piccinelli2024unidepth}, illustrating that our model is agnostic to the choice of depth priors. Using estimated camera poses, we back-project points into 3D via refined depth maps, yielding accurate camera motion estimation and consistent reconstructions. Our method remains robust in challenging scenarios with multiple moving people, rapid egocentric camera motion, and dynamic object interactions.

\begin{figure}[t]
    \centering
    \includegraphics[width=\linewidth]{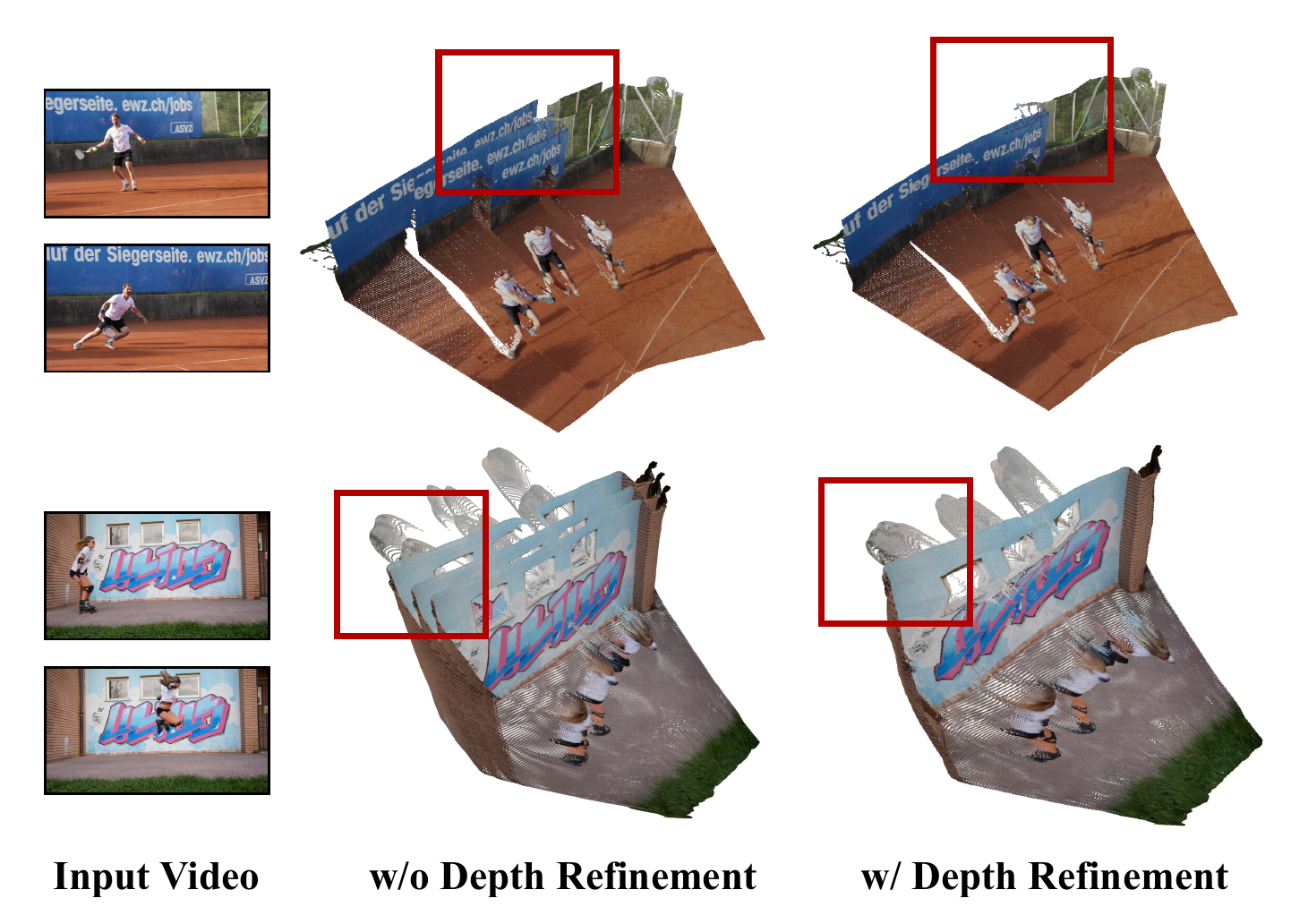}
    \vspace{-2em}
    \caption{\textbf{Visualization of global refinement on the DAVIS~\cite{khoreva2019video}.} We compare the 3D consistency of fused point clouds obtained from raw depth priors and our refined depths. Our refined depths yield a significantly more consistent 3D structure over time.}%
    \label{fig:depth_refinement}
    \vspace{-0.8em}
\end{figure}

The superior reconstruction quality primarily stems from the proposed global refinement. To demonstrate this, we conduct an ablation study as shown in \cref{tab:depth_ablation}.
We experiment with switching off $\mathcal{L}_\text{depth}$ and $\mathcal{L}_\text{rigid}$---defined in \cref{eq:loss_depth} and \cref{eq:loss_rigid}---together, as well as alternately.
Omitting both losses leads to suboptimal accuracy.
Including $\mathcal{L}_\text{depth}$ alone reduces the Abs Rel error from 0.121 to 0.103 and boosts the inlier percentage from 89.6\% to 95.0\%.
The rigid consistency alone offers modest gains, but the best result is achieved by combining both losses, illustrating their mutual complementarity. 
\cref{fig:depth_refinement} shows two intuitive comparisons between the reconstructions with and without our depth refinement. Directly fusing the monocular depth map with the estimated camera pose leads to inconsistent reconstructions with duplicated 3D structures, whereas our global refinement significantly improves 3D coherence. 

\section{Conclusion}

In this work, we addressed the challenge of reconstructing dynamic scenes from casual video sequences using a traditional SLAM method. We equipped a 3D point tracker with a motion decoupling mechanism to separate camera-induced motion from object motion.
Operating on the isolated static motion components, bundle adjustment leads to substantial improvements in camera pose accuracy and reconstruction quality.
Additionally, global refinement leverages the sparse depth estimates from BA to ensure scale consistency and alignment of depth maps across video frames.
Overall, \oursName enables precise 3D trajectory estimation and dense reconstructions in highly dynamic environments.
More broadly, our work demonstrates that traditional optimization enhanced by deep priors can provide an accurate and robust solution to challenging real-world scenarios.

{\small
\inparagraph{Acknowledgements.}
This work was supported by the ERC Advanced Grant SIMULACRON, by the Federal Ministry for the Environment, Nature Conservation, Nuclear Safety and Consumer Protection (BMUV) through the AuSeSol-AI project (grant 67KI21007A), and by the TUM Georg Nemetschek Institute Artificial Intelligence for the Built World (GNI) through the AICC project.}
\\
{
    \small
    \bibliographystyle{ieeenat_fullname}
    \bibliography{main}
}

\appendix
\clearpage
\maketitlesupplementary

\pagenumbering{roman}

\section{Implementation Details}%
\label{sec:impl}

\inparagraph{Model Architecture.}
We develop our 3D tracker based on CoTracker~\cite{karaev2023cotracker}, incorporating additional modifications.
The feature extractor is a CNN-based architecture, consisting of a $7 \times 7$ convolutional layer followed by several $3 \times 3$ residual blocks, which generate multi-scale feature maps.
These feature maps are aggregated into a single feature map with additional convolutional layers, producing a resolution of $\frac{1}{4}$ of the input image.
We adopt the depth map encoding strategy from SpatialTracker~\cite{xiao2024spatialtracker}, omitting the tri-plane correlation for improved efficiency.

The 3D tracker $\mathcal{T}$ is a transformer-based architecture that alternates between spatial and temporal attention blocks, consisting of 6 layers.
For motion decoupling, we employ a smaller 3-layer transformer  $\mathcal{T}_{\text{dyn}}$ designed specifically to predict the static component.
Each layer in both transformers includes a temporal and a spatial attention block, with each block comprising an attention layer followed by an MLP\@.

\inparagraph{Refinement Formulation.}
The 3D tracker uses an iterative transformer-based refiner module $\mathcal{T}$~\cite{karaev2023cotracker}.
To provide the initial state as input to $\mathcal{T}$, we copy the 2D location of the query point across all frames $s \in (1, \dots, S)$ and sample the point features as
\begin{equation}
    X^{(0)}(s) = X, \qquad f^{(0)}(s) = F_t[\mathbf{x}],
\end{equation}
where the superscript $^{(\cdot)}$ denotes the iteration index.
Aggregating context over all timesteps in the window, the 3D tracker $\mathcal{T}$ iteratively updates the point features $f(s)$ and the 3D trajectories $X(s)$.
Dropping timestep $s$ to avoid clutter, the update in the $k$-th iteration is
\begin{equation}
    (X^{(k+1)}, \: f^{(k+1)}) = \mathcal{T}(f^{(k)}, \text{PE}(X^{(k)}), C^{(k)}),
\end{equation}
where $\text{PE}(\cdot)$ represents the positional embedding of the point track, encoding its 3D location and timestep with periodic bases.

For the dynamic tracker, $\mathcal{T}_\text{dyn}$ predicts the object-induced motion $X_{\text{dyn}}$ and dynamic point feature $f_{\text{dyn}}$ as
\begin{equation}
(X_{\text{dyn}}^{(k+1)}, f_{\text{dyn}}^{(k+1)}) = \mathcal{T}_{\text{dyn}}(f_{\text{dyn}}^{(k)}, \text{PE}(X_{\text{static}}^{(k)}), C^{(k)}),
\end{equation}
where $X_{\text{dyn}}^{(0)}$ is set to 0 and $f_{\text{dyn}}^{(0)}$ is set to $F_t[\mathbf{x}]$.

\inparagraph{Local Context.}
The local correlation map $C^{(k)}$ is a function of the point track $X^{(k)}$, \ie
\begin{equation}
    C^{(k)}(s) = C[X^{(k)}(s)],
\end{equation}
and serves to enhance the spatial context of each tracked point.
To compute $C$, we follow~\cite{xiao2024spatialtracker} and first apply Fourier embedding to the depth map $\mathbf{D}$ to obtain depth features $\mathbf{D}^{\text{Fourier}}$.
These are concatenated with the image features $\mathbf{F}$ to produce fused features $\mathbf{F}^{\text{hyb}}$.
We then apply bilinear interpolation to generate multi-scale hybrid feature maps $\Tilde{\mathbf{F}}^{\text{hyb}}$.
For each point $X^{(k)}(s)$, we extract a local region $\Tilde{\mathbf{F}}^{\text{hyb}}_s$ centered at $X^{(k)}(s)$, and compute the correlation as the inner product between the point feature $f^{(k)}(s)$ and the surrounding hybrid features~\cite{karaev2023cotracker}.
The resulting correlation map $C^{(k)}$ captures both appearance and geometric cues in a unified representation.

\inparagraph{Training.}
We train our model on the TAP-Vid-Kubric training set~\cite{doersch2022tap}, which includes 11,000 sequences.
Each sequence consists of 24 frames derived from the MOVi-F dataset.
Following the data preprocessing steps from CoTracker~\cite{karaev2023cotracker}, we additionally extract dynamic labels, 3D total trajectory, and static trajectory ground truth.
The static trajectory ground truth is generated by back-projecting queries from their 3D positions in the source frame into the target frames using ground-truth camera poses.
The original image resolution of each sequence is $512 \times 512$, and we crop it to $384 \times 512$ during training.
For image augmentation, we apply random resizing, flipping, cropping, Gaussian blurring, and color jitter.
For depth augmentation, we adopt the scale-shift augmentation and Gaussian blurring techniques described in~\cite{gao2018ldso}.
The augmented ground-truth depth is consistently used during the training process.

\inparagraph{Bundle Adjustment.}
We build our bundle adjustment framework based on DPVO~\cite{teed2024deep} and extend it to support RGB-D bundle adjustment.
To enhance the robustness of pose estimation, we incorporate weight filtering during the pose update computation.
We set the visibility threshold $\delta_v = 0.9$ and the dynamic label threshold $\delta_m = 0.9$ to ensure that camera pose updates rely exclusively on reliable point trajectories.
This approach is adopted because recovering the camera pose primarily depends on a few accurate correspondences.
In contrast, for depth updates, we consider all point trajectories to fully leverage the static components estimated through motion decoupling.

\section{Additional Ablation Experiments}

\begin{figure}[t]
\centering
\includegraphics[width=1.0\linewidth]{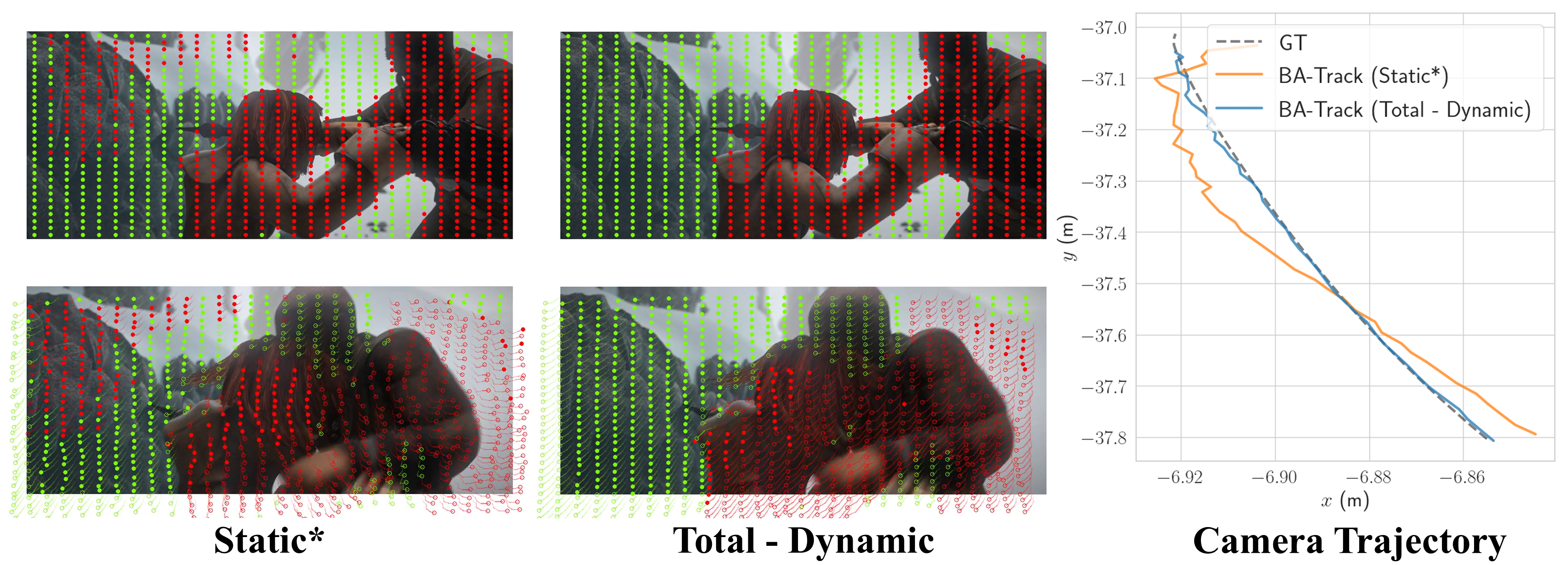}
\caption{\textbf{Comparison of different trackers.} Red: estimated dynamic point tracks.
Green: estimated static point tracks.
The resulting camera trajectories from different trackers are shown on the right.
Our dual networks with motion decoupling yield more accurate camera poses.}%
\label{fig:ablation_tracker}
\end{figure}

\begin{figure}[ht]
\centering
\includegraphics[width=\linewidth]{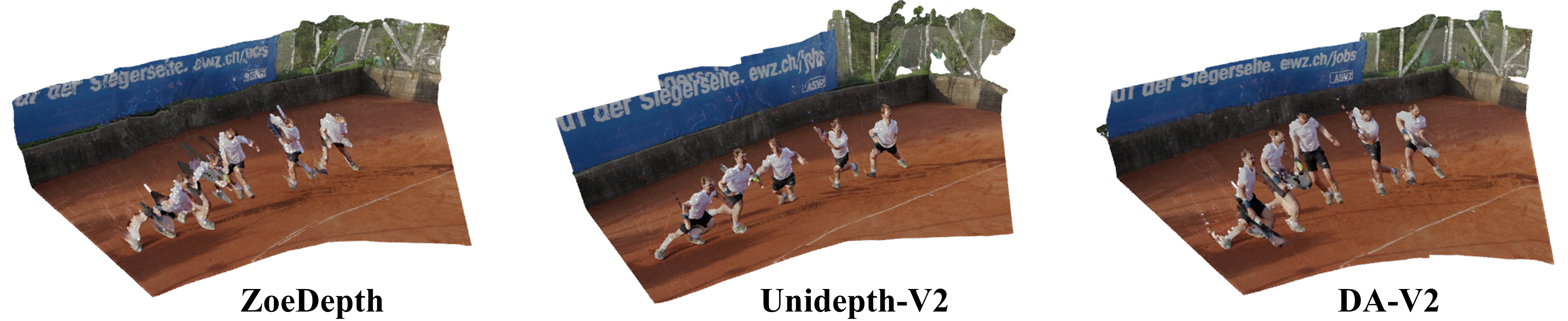}
\caption{\textbf{Qualitative comparison across different depth priors.} Our method is robust to different types of depth priors.}%
\label{fig:ablation_depth_priors}
\end{figure}

\inparagraph{Comparison of single and dual network architectures.}
In \cref{fig:ablation_tracker}, we further compare \textit{(1)} a single-network tracker predicting only the static motion (Static*) and \textit{(2)} our default tracker predicting the total and dynamic components.
Static* struggles to capture the camera motion for dynamic points and produces motion label outliers, degrading pose estimates.
This is likely because Static* predicts similar camera-induced motion for all points, making dynamic tracks hard to distinguish.

\inparagraph{Robustness with different depth priors.}
To evaluate the robustness of our method to different depth prior models, we use two additional depth backbones: UniDepth-V2~\cite{piccinelli2024unidepth} and Depth‑Anything-V2 (DA‑V2)~\cite{piccinelli2025unidepthv2}, with per-video metric alignment based on the UniDepth-V2 scale.
We compare the dynamic reconstruction results of our method using ZoeDepth, UniDepth-V2, and DA‑V2.
As shown in \cref{fig:ablation_depth_priors}, our method remains robust across different depth priors for both camera pose estimation and reconstruction.

\inparagraph{Failure modes.}
Our motion-decoupled tracker is robust to short-term occlusions and moderate motion but may fail under prolonged occlusions or complex, unseen motions.
While BA-Track generalizes well to diverse rigid motion and performs strongly on most crowded scenes in Shibuya and DAVIS, its accuracy can degrade under severe occlusion or when tracking many small objects.
Additionally, the method struggles with highly deformable or non-rigid objects, which are underrepresented in the TAP-Vid-Kubric~\cite{doersch2022tap} training set.
Expanding training to larger and more diverse dynamic datasets could address these limitations.

\section{Additional Qualitative Results}%
\label{sec:vis}

\begin{figure*}[t]
    \centering
    \includegraphics[width=\linewidth]{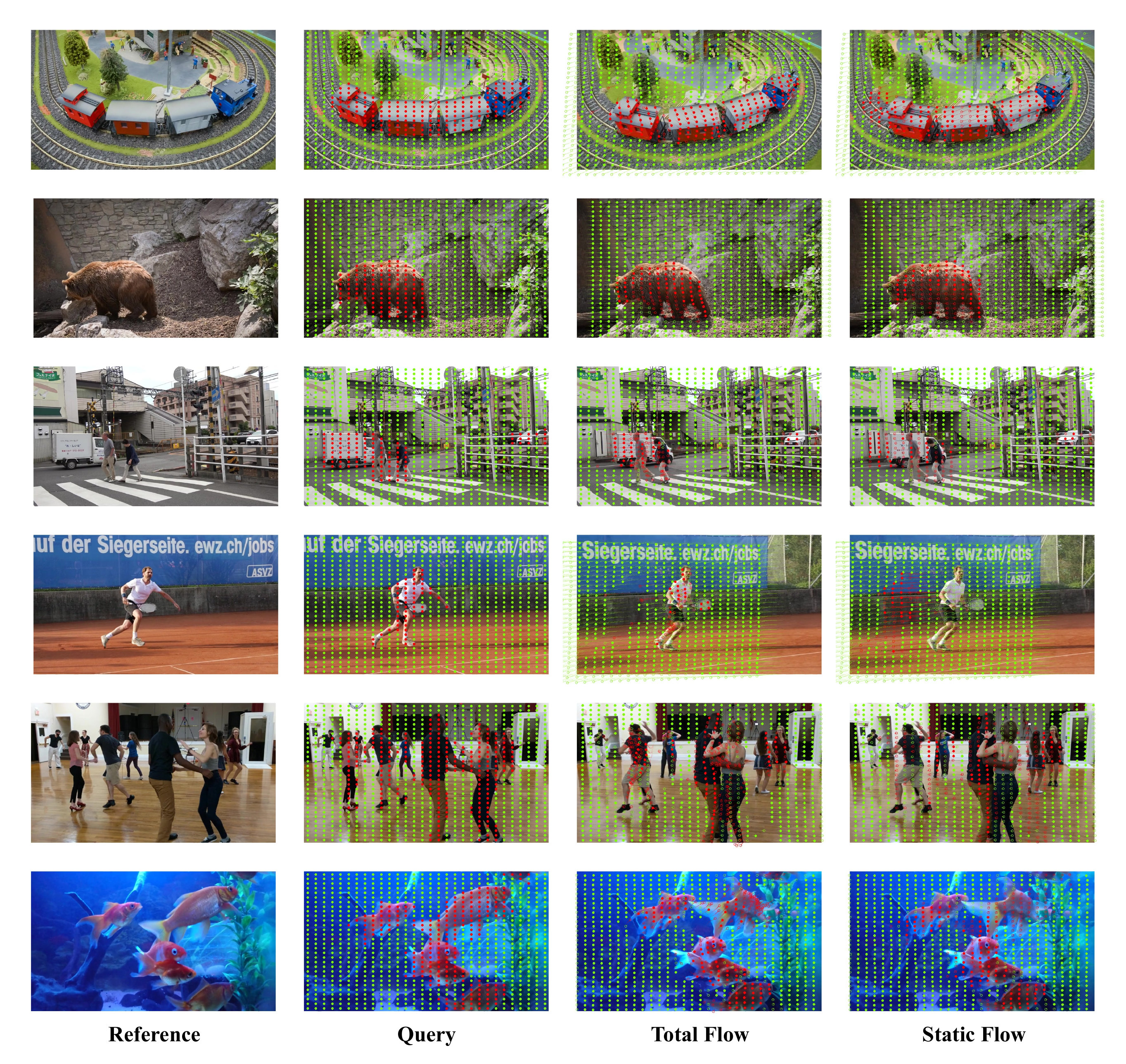}
    \vspace{-2em}
    \caption{\textbf{Visualization of motion decoupling on the DAVIS dataset~\cite{khoreva2019video}.} The reference frame corresponds to the first frame of the video, from which the queries are extracted.
    The total flow represents the combined motion of points, while the static flow refers to the motion attributed solely to its static components.
    The estimated static and dynamic trajectories are depicted in green and red, respectively.
    We observe that the red points in the static flow largely remain in their original reference frame, signifying successful motion decoupling.}%
    \label{fig:supp_motion_decouple}
\end{figure*}

\inparagraph{Motion Decoupling.}
We present additional results for motion decoupling on various video samples from the DAVIS~\cite{khoreva2019video} dataset, as shown in~\cref{fig:supp_motion_decouple}.
The examples cover different motion scenarios, including single-object motion, multi-object motion, and occlusions.
Our motion decoupling point tracker effectively distinguishes dynamic trajectories from static trajectories, providing robust estimates of the static components (illustrated in the last column).
Treating the dynamic parts as static enables us to incorporate them into the geometry recovery process through bundle adjustment.

\section{Limitations and Discussion}%
\label{sec:disc_and_future}

\inparagraph{Joint Refinement with Camera Parameters.}
By design, BA-Track assumes the camera intrinsic parameters are provided either as ground truth or as estimates from off-the-shelf methods.
These parameters are essential for bundle adjustment, which recovers the camera pose and the sparse point depths.
However, calibration errors or estimation noise can introduce inaccuracies in the intrinsic parameters and lead to unreliable reprojection loss.
One way to address this limitation is to extend our framework to jointly optimize camera intrinsics along with pose and point depths.
Starting from an initial estimate, the intrinsics can be iteratively refined during optimization~\cite{schonberger2016structure}.
Future work could integrate this refinement with better initialization to improve convergence and reduce computational cost.

\inparagraph{Depth Refinement.}
To enable track-guided depth refinement, we introduce a deformable scale map applied to the original dense depth map.
This scale map facilitates refinement by aggregating information from sparse point tracks in a smooth and coherent manner, bridging the gap between sparse tracks and dense depth mapping.
While effective, the scale map has limitations in handling complex error patterns present in monocular depth estimates.
Future work could explore refinement models based on dense vector fields, which may improve continuity and smoothness.
Additionally, representing the depth map with neural networks and refining it by backpropagating gradients to update the network weights presents another promising avenue for exploration~\cite{zhang2022structure}.

\end{document}